\documentclass[letterpaper, 10 pt, conference]{ieeeconf}  

\IEEEoverridecommandlockouts                              

\usepackage[table,xcdraw]{xcolor}
\usepackage{etoolbox} 
\usepackage[utf8]{inputenc} 
\usepackage[T1]{fontenc}    
\usepackage[english]{babel} 
\usepackage[protrusion=true,expansion=true]{microtype}
\usepackage{comment}
\usepackage{cancel}
\usepackage{csquotes}
\usepackage{listings}
\usepackage{nameref}
\usepackage{paralist}
\usepackage{xspace}
\usepackage{makeidx}
\usepackage{pdfpages}
\usepackage{footnote}

\usepackage{hyperref}
\hypersetup{
    colorlinks=true,
    linkcolor=black,
    citecolor=black,
    filecolor=cyan,
    urlcolor=blue
}

\usepackage{flushend}
\usepackage{balance}

\usepackage{paralist}
\let\labelindent\relax 
\usepackage{enumitem}
\setlist[itemize]{noitemsep,leftmargin=*,topsep=0in}
\setlist[enumerate]{noitemsep,leftmargin=*,topsep=0in}

\usepackage{amsmath,amssymb} 
\usepackage{amsfonts}       
\usepackage{mathtools}
\usepackage{array} 
\usepackage{nicefrac}       
\usepackage{breqn}          
\usepackage{textcomp}
\usepackage{bm} 
\usepackage{pifont}
\usepackage[
  separate-uncertainty = true,
  multi-part-units = repeat
]{siunitx}

\usepackage{graphicx}
\usepackage{adjustbox} 
\usepackage{epsfig}

\usepackage{float}
\usepackage[font=small]{subcaption}
\usepackage[font=small,labelfont=bf]{caption}
\usepackage{lscape}      
\usepackage{wrapfig}

\usepackage{tikz}  
\usepackage{makecell}
\usepackage{overpic}
\usepackage{algorithm}
\usepackage[noend]{algpseudocode}

\usepackage{tabularx}
\usepackage{multirow}
\usepackage{multicol}
\usepackage{booktabs}   
\usepackage{array} 

\usepackage{longtable}
\usepackage{threeparttable}

\makesavenoteenv{tabular}
\makesavenoteenv{table}

\usepackage{cite}

\renewcommand{\paragraph}[1]{\vspace{0.2em}\noindent\textit{#1} --}

\newcommand\mybar{\kern1pt\rule[-\dp\strutbox]{.8pt}{\baselineskip}\kern1pt}

\newcommand{\sysName}{\textsc{SoftMimicGen}\xspace}
\newcommand{\MimicGen}{\textsc{MimicGen}\xspace}

\title{\LARGE \bf
\sysName: A Data Generation System for Scalable Robot Learning in Deformable Object Manipulation
}

\author{
Masoud Moghani$^{1,2}$,
Mahdi Azizian$^{1}$,
Animesh Garg$^{3}$,
Yuke Zhu$^{1}$,
Sean Huver$^{*,1}$,
Ajay Mandlekar$^{*,1}$%
\thanks{$^{1}$NVIDIA, $^{2}$University of Toronto, $^{3}$Georgia Institute of Technology}%
\thanks{*Equal Advising, Correspondence to: \href{mailto:moghani@cs.toronto.edu}{moghani@cs.toronto.edu}}%
}

\let\oldtwocolumn\twocolumn
\renewcommand\twocolumn[1][]{%
    \oldtwocolumn[{#1}{
    \begin{center}
    \vspace{-.34in}
    \includegraphics[width=0.95\linewidth]{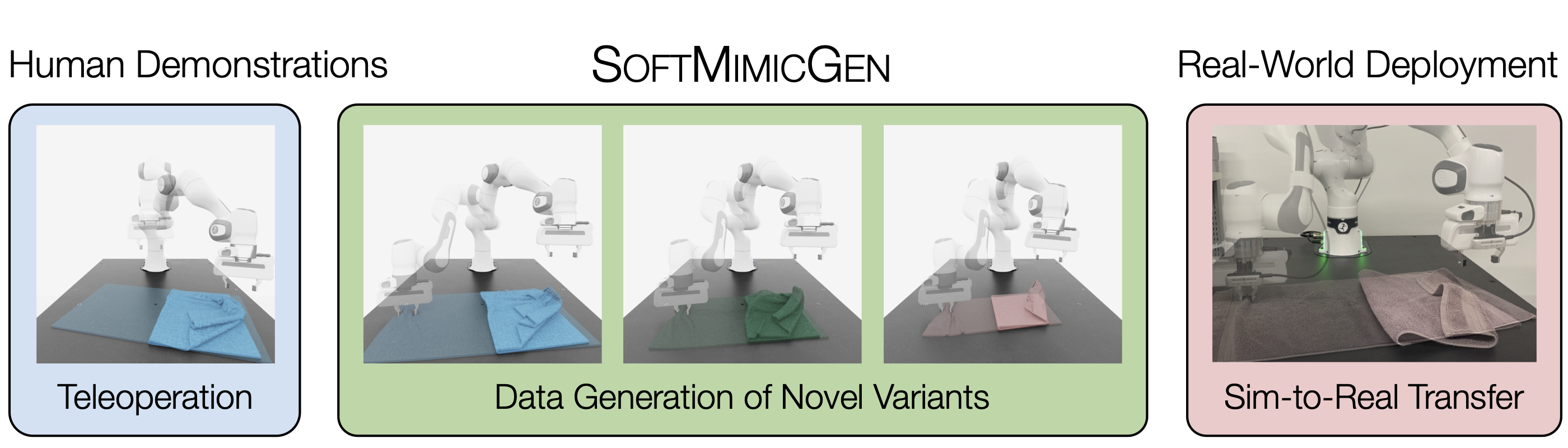}
    \captionof{figure}{\textbf{\sysName Overview.}
    \sysName provides an efficient pipeline for synthesizing robot trajectories for deformable object manipulation. (Left) A human teleoperator first collects a small set of fine-grained, dexterous robot trajectories. (Center) \sysName then generates large-scale datasets for novel instances of deformable objects across new contexts. (Right) Generated demonstrations enable the training of high-performing policies in simulation, which can then be transferred to real-world platforms. \sysName is compatible with diverse robot embodiments and tasks that require dynamic, contact-rich manipulation.}
    \label{fig:fig1}
    \end{center}
    }]
}

\begin{document}

\maketitle
\thispagestyle{empty}
\pagestyle{empty}

\begin{abstract}

Large-scale robot datasets have facilitated the learning of a wide range of robot manipulation skills, but these datasets remain difficult to collect and scale further, owing to the intractable amount of human time, effort, and cost required. Simulation and synthetic data generation have proven to be an effective alternative to fuel this need for data, especially with the advent of recent work showing that such synthetic datasets can dramatically reduce real-world data requirements and facilitate generalization to novel scenarios unseen in real-world demonstrations. However, this paradigm has been limited to rigid-body tasks, which are easy to simulate. Deformable object manipulation encompasses a large portion of real-world manipulation and remains a crucial gap to address towards increasing adoption of the synthetic simulation data paradigm. In this paper, we introduce \sysName, an automated data generation pipeline for deformable object manipulation tasks. We introduce a suite of high-fidelity simulation environments that encompasses a wide range of deformable objects (stuffed animal, rope, tissue, towel) and manipulation behaviors (high-precision threading, dynamic whipping, folding, pick-and-place), across four robot embodiments: a single-arm manipulator, bimanual arms, a humanoid, and a surgical robot. We apply \sysName to generate datasets across the task suite, train high-performing policies from the data, and systematically analyze the data generation system. Project website: \href{https://softmimicgen.github.io}{softmimicgen.github.io}.

\end{abstract}

\section{Introduction}

Robot foundation models~\cite{brohan2023rt, kim2024openvla, black2024pi_0, bjorck2025gr00t}, trained on a combination of web-scale vision-language data and large-scale robot manipulation datasets, have shown an impressive capability to perform a wide range of complex manipulation tasks autonomously. However, these advances have been fueled chiefly by the availability of large robot manipulation datasets. These datasets are often collected via robot teleoperation by large teams of human operators over several time-consuming months~\cite{ebert2021bridge, brohan2022rt, o2024open, khazatsky2024droid}. While the robot teleoperation paradigm provides a simple means towards collecting robotics datasets, it remains a costly and labor-intensive endeavor, and hinders the broader development of robot foundation models.

Simulation is a promising alternative to fuel the need for large robot manipulation datasets, especially due to several recent developments. Designing high-quality simulation environments is becoming easier, due to the availability of high-fidelity physics simulators and photorealistic rendering~\cite{todorov2012mujoco, mittal2025isaac}, and the advent of generative AI tools, which facilitate the automated generation of scenes, assets, and tasks~\cite{wang2023robogen, robocasa2024}. Recent automated data generation tools make it possible to synthesize large amounts of diverse, high-quality robot manipulation demonstrations with little human effort~\cite{dalal2023imitating, mandlekar2023mimicgen, jiang2024dexmimicgen, garrett2024skillmimicgen}. Furthermore, recent work highlights that large-scale synthetic simulation datasets can easily be used to train high-performance real-world manipulation policies by \textit{co-training} on these synthetic simulation datasets and small amounts of real-world data~\cite{maddukuri2025sim, wei2025empirical, bjorck2025gr00t}. This synthetic data generation paradigm can drastically reduce real-world data requirements and facilitate generalization to novel scenarios unseen in the real-world datasets. However, this paradigm has been useful only for tasks that can be easily simulated, which has limited its application to mostly rigid-body manipulation tasks.

Deformable object manipulation encompasses a significant portion of real-world manipulation and remains a crucial gap towards increasing the adoption of simulation tools and synthetic data generation. However, overcoming this gap is challenging for several reasons. First, simulating deformable objects is a difficult problem computationally, and getting interactions with such objects to simulate in real-time (or faster) is even more challenging. Sourcing and annotating simulation assets for deformable manipulation tasks such that they behave as anticipated is also non-trivial. Second, common synthetic data generation solutions assume that objects can be assigned a rigid frame of reference, and exploit robot manipulation motion invariance with respect to this frame in order to generate new demonstrations~\cite{mandlekar2023mimicgen, jiang2024dexmimicgen, garrett2024skillmimicgen}. However, this assumption breaks down quickly for deformable objects, since they do not remain rigid, requiring new algorithmic considerations for data generation~\cite{schulman2016learning, schulman2013case, lee2014unifying}.

\textbf{Towards addressing these challenges, we create a simulation suite of deformable object manipulation tasks, and develop \sysName, a synthetic data generation pipeline for deformable object manipulation.} The simulation suite leverages recent advancements in deformable object simulation, ensuring all environments simulate in real-time (or faster). \sysName builds upon \MimicGen~\cite{mandlekar2023mimicgen}, which is an object-centric trajectory generation system for rigid object manipulation. Like \MimicGen, \sysName starts with a small source set of human demonstrations (1 to 10) and generates much larger datasets automatically. The key component of \MimicGen is to extract and replay object-centric demonstrations from the source dataset by leveraging a static object reference frame that remains constant between the source dataset and the new task instance. However, no such static reference frame is guaranteed to exist for deformable objects -- consequently, \sysName instead leverages non-rigid registration techniques~\cite{schulman2016learning, schulman2013case} to adaptively transform source demonstrations while accounting for the changed state of deformable objects. This results in a more capable demonstration generation mechanism that can be applied to deformable and rigid object tasks alike. We train visuomotor policies via imitation learning on generated data, enabling direct manipulation of deformable objects from images without explicit registration at inference time.

\noindent \textbf{Summary of contributions:}
\newline\noindent $\bullet$ \sysName enables synthetic data generation for deformable object manipulation tasks.
\newline\noindent $\bullet$ We release a suite of high-fidelity simulation environments that encompass a range of deformable objects (stuffed animal, rope, tissue, towel) and manipulation behaviors (high-precision threading, dynamic whipping, folding, pick-and-place) across four robot embodiments (Franka and YAM robot arms, humanoid, surgical robot).
\newline\noindent $\bullet$ We apply \sysName to generate thousands of demonstrations per task, train high-performing policies from the data, and provide a systematic analysis of the data generation system.
\newline\noindent $\bullet$ We demonstrate real-world deployment across diverse deformable manipulation tasks, where policies trained on \sysName-generated data achieve zero-shot sim-to-real transfer and are further improved via sim-real co-training.

\section{Related Work}

\textbf{Deformable Object Manipulation.}
Research in the robotic manipulation of deformable objects, including cloth, granular media, and soft materials, has pivoted from classical physics-based models to data-driven, learning-centric paradigms~\cite{zhu2022domsurvey,yin2021modeling}. Multiple methods construct explicit simulations using pre-scanned static objects and point cloud observations. Most recent approaches build upon SDFs~\cite{qiao2022neuphysics}, NeRF~\cite{feng2024pie}, or Gaussian Splatting~\cite{zhong2024reconstruction} to support flexible physical digital twin creation. Furthermore, neural methods of dynamics learning using graph-based representations have been used to learn the dynamics of various types of deformable objects such as plasticine~\cite{shi2023robocook}, cloth~\cite{li2020causal}, and fluid~\cite{sanchez2020learning}. Yet, studying deformable object manipulation in the context of large-scale imitation learning and foundation models remains a challenge due to the scarcity of open-source simulation environments and datasets. \sysName is a first step towards enabling this kind of investigation.

\textbf{Data Collection and Data Generation for Robotics.}
Robot teleoperation~\cite{mandlekar2018roboturk, zhao2023learning} is a popular option for collecting demonstrations to train robots -- humans use a teleoperation device (such as a smartphone or VR controller) to control a robot and perform different manipulation tasks. The robot sensor streams and controller actions are recorded into a dataset. This paradigm has been scaled up extensively in recent years through the use of teams of human operators and robots over extended periods of time~\cite{ebert2021bridge, brohan2022rt, o2024open, khazatsky2024droid}. Other works have used pre-programmed demonstrators~\cite{james2020rlbench, zeng2020transporter, wang2023robogen, dalal2023imitating}, but scaling these approaches to a larger variety of tasks can be difficult. The size of collected datasets can be increased using offline data augmentation~\cite{kostrikov2020image, young2020visual, mandlekar2021matters}. These include the use of generative models to augment observations~\cite{yu2023scaling, chen2023genaug}, and leveraging counterfactual reasoning to augment observation-action pairs~\cite{pitis2020counterfactual, pitis2022mocoda}.

\MimicGen~\cite{mandlekar2023mimicgen} is a data generation framework that exploits object-centric invariance to generate new trajectories. DexMimicGen~\cite{jiang2024dexmimicgen} extended this approach to bimanual manipulation, and SkillMimicGen~\cite{garrett2024skillmimicgen} extended the approach to incorporate motion planning (complementary to our approach). \sysName, like \MimicGen, generates new datasets online, but can be applied to a much broader set of tasks by using an improved trajectory transformation process based on non-rigid registration techniques.

\textbf{Learning Manipulation from Human Demonstrations.}
Behavioral Cloning (BC)~\cite{pomerleau1989alvinn} is a widely used approach for learning robot manipulation policies from demonstrations~\cite{schaal1999imitation, Ijspeert2002MovementIW, mandlekar2021matters, chi2023diffusion}. It consists of training the agent to produce actions that are consistent with observation-action pairs in the training dataset. This method has been shown to be extremely effective for robot manipulation~\cite{Billard2008RobotPB, Calinon2010LearningAR, zeng2020transporter, black2024pi_0, khazatsky2024droid, brohan2023rt}, but the quality of the results depends on the availability of large-scale, high-quality manipulation datasets. Some recent works~\cite{maddukuri2025sim, wei2025empirical, bjorck2025gr00t} have shown that synthetic simulation data can supplement real-world datasets and reduce the amount of costly real-world data that is required.

\section{Prerequisites}

\subsection{Behavioral Cloning}
\label{subsec:bc}

We model each manipulation task as a Partially Observable Markov Decision Process (POMDP). We are given a dataset of $N$ demonstrations $\mathcal{D} = \{(s_0^i, o_0^i, a_0^i, s_1^i, o_1^i, a_1^i, ..., s_{H_i}^i)\}_{i=1}^N$ with states $s \in {\cal S}$, observations $o \in {\cal O}$, and actions $a \in {\cal A}$. Each episode starts in an initial state $s_0^i \sim D$ sampled from the initial state distribution $D \subseteq {\cal S}$. The goal is to learn a policy $\pi: {\cal O} \to {\cal A}$ that takes observations as inputs and outputs a distribution over the action space. Policies are trained using Behavioral Cloning~\cite{pomerleau1989alvinn} via the maximum likelihood objective $\arg\max_{\theta} \mathbb{E}_{(s, o, a) \sim \mathcal{D}} [\log \pi_{\theta}(a \mid o)]$ using datasets generated by \sysName.

\subsection{Problem Statement}
\label{subsec:problem}

We are given a source dataset $\mathcal{D}_{\text{src}}$ consisting of a small (typically 1 to 10) number of human demonstrations, and our goal is to use it to generate a large dataset of demonstrations $\mathcal{D}$ on either the same task, or a task with a different initial state distribution $D' \subseteq {\cal S}$ (typically one with a larger set of possible placements for objects in the scene). To generate a new demonstration, (1) a start state is sampled from $D'$, (2) one or more demonstrations $\tau \in \mathcal{D}_{\text{src}}$ are selected and adapted to produce and execute a new robot trajectory $\tau'$, and (3) if the task is completed successfully, the trajectory is added to the generated dataset. The core problem that must be addressed is the mechanism used to carry out step (2) -- namely, the trajectory generation.

\subsection{Assumptions}
\label{subsec:assumptions}

We make the following assumption on how deformable objects are represented: \textbf{(A0):} every deformable object is represented as a collection of 3-dimensional node positions, $O = \{\mathbf{n_i}\}_{i=1}^{N_{O}}$, where $\mathbf{n_i} \in \mathbb{R}^3$ are 3-dimensional positions $(x_i, y_i, z_i)$, and $N_O$ is the number of nodes for the object. These node positions are typically obtained through a simulator's soft object solver. Note that this is equivalent to a point cloud representation for objects, and rigid-body objects could also be represented in this manner.

We also make additional assumptions, similar to \MimicGen~\cite{mandlekar2023mimicgen}, namely: \textbf{(A1):} The action space $\mathcal{A}$ consists of pose commands for an end-effector controller for each robot arm, and a gripper command for each arm. \textbf{(A2):} Each task can be divided into sequences of object-centric subtasks $(S_1(o_{S_1}), S_2(o_{S_2}), ..., S_M(o_{S_M}))$ for each arm, where the arm primarily interacts with one object. However, unlike \MimicGen, the manipulation need not be relative to a specific object coordinate frame, which can be ill-posed for deformable objects. \textbf{(A3):} During data collection, we assume that object configurations can be observed or estimated prior to the start of each subtask. For rigid objects, this corresponds to the object pose; for deformable objects, it corresponds to the positions of all object nodes (see \textbf{A0}).

\subsection{\MimicGen}
\label{subsec:mimicgen}

\MimicGen first parses the source demonstrations in $\mathcal{D}_{\text{src}}$ into contiguous object-centric manipulation segments $\{\tau_i\}_{i=1}^M$, each of which corresponds to a subtask $S_i(o_i)$. Each of these segments is a sequence of end-effector control poses $\tau_i = (T^{C_0}_W, T^{C_1}_W, ..., T^{C_K}_W)$ where $W$ is the world reference frame. The segmentation can be done autonomously with heuristics or using human annotation. To generate a demonstration in a novel scene, it first observes the pose of the object for the current subtask $T^{o'_i}_{W}$. It then transforms the poses in a source human segment such that the relative poses between the end-effector frame and object frame are preserved in both the source segment and the new scene -- this is the invariance that \MimicGen exploits. \MimicGen carries out this transformation using a constant SE(3) transform $T^{o'_i}_{W}(T^{o_i}_{W})^{-1}$. \MimicGen then adds poses to the start of the transformed segment in order to linearly interpolate between the robot's current pose and the start of the transformed segment (which may start far from the current pose). It then executes the sequence of poses in the segment and repeats the process for the next subtask until all subtasks have been executed, and keeps the executed trajectory as a demonstration if it was successful. Unlike \MimicGen, \sysName must deal with deformable objects with nonlinear material properties and elasticity -- consequently, it is not straightforward to use the same rigid SE(3) transformation strategy to exploit object-centric invariance. \sysName overcomes this challenge by estimating non-rigid transformations between deformable object states (expressed as collections of node positions, see Assumption \textbf{A0} in Sec.~\ref{subsec:assumptions}) and leveraging these transformations to perform non-rigid spatial adaptation of trajectory segments.

\begin{figure*}
    \centering
    \includegraphics[width=0.82\textwidth]{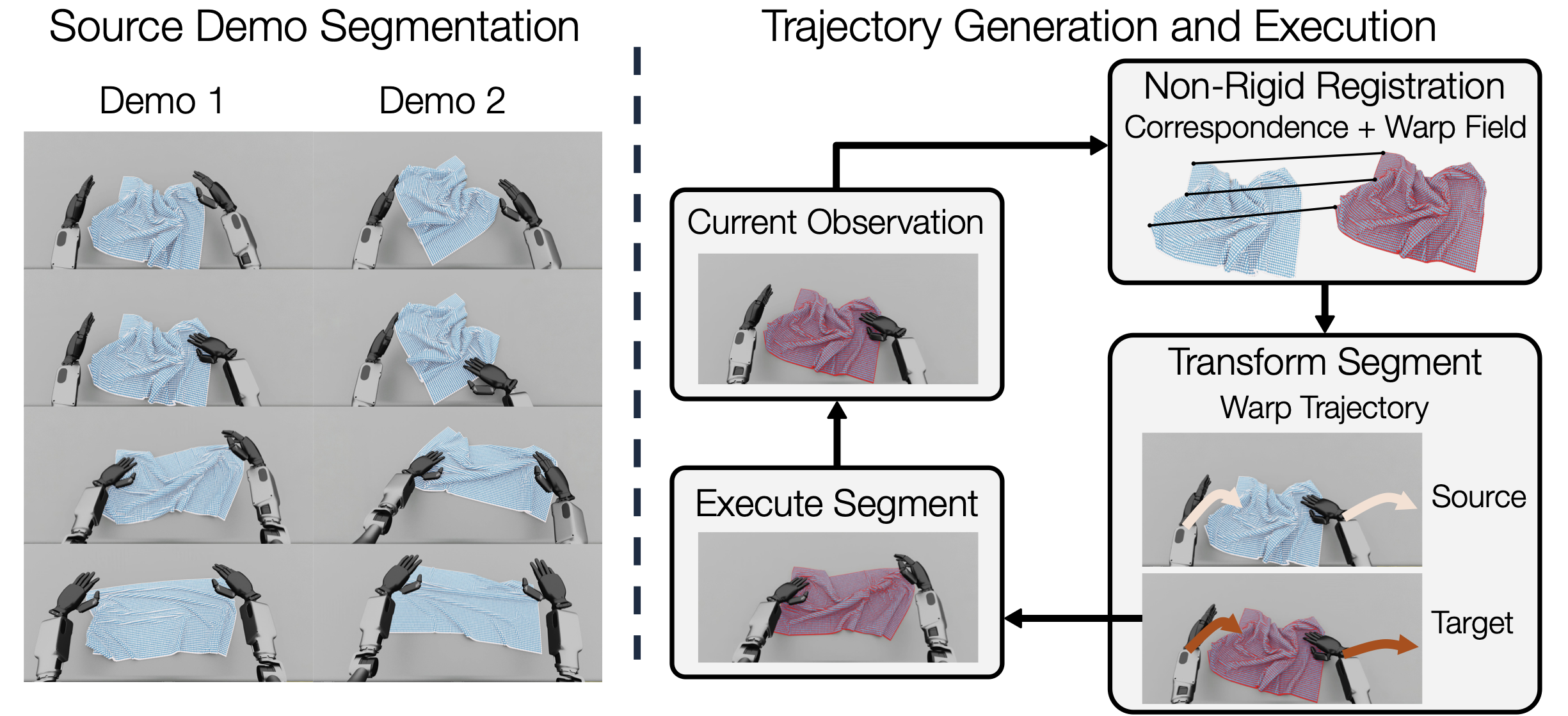}
    \caption{\textbf{\sysName\ System Pipeline.}
    \textit{(Left)} Human-teleoperated demonstrations are segmented into object-centric subtasks using manual annotations or heuristic signals, forming a library of source segments.
    \textit{(Right)} Given a new target scene, \sysName\ \textit{(i)}~observes the current deformable state, \textit{(ii)}~performs non-rigid registration to establish correspondence and a warp field between the source and target geometries (\textit{correspondence + warp field}), \textit{(iii)}~selects the source segment with the lowest registration cost and applies the resulting warp field to the end-effector trajectory (\textit{warp trajectory; source end-effector trajectory in light arrows, warped trajectory in darker arrows}), and \textit{(iv)}~executes the warped trajectory in the target environment.}
    \label{fig:fig2}
\end{figure*}

\section{\sysName}

\sysName enables large-scale data generation for deformable object manipulation using only a small number of human teleoperated demonstrations. As described in Sec.~\ref{subsec:problem}, to generate a new demonstration, reference source demonstrations must be selected, and adapted appropriately. As in \MimicGen, we seek to exploit an object-centric invariance to re-purpose existing source human demonstrations and generate new trajectories, but doing this for deformable objects is not as straightforward. We first describe how deformable objects are represented and why the \MimicGen strategy does not suffice (Sec.~\ref{subsec:deformable}). Next, we describe how non-rigid registration can offer a solution to the problem of object-centric trajectory transfer for deformable objects (Sec.~\ref{subsec:nonrigid}). Finally, we describe how we incorporate this mechanism in the data generation process (Sec.~\ref{subsec:datagen_mechanism}).

\subsection{Deformable Object Representation}
\label{subsec:deformable}

We assume that every deformable object is described by a collection of 3-dimensional node positions, $O = \{\mathbf{n_i}\}_{i=1}^{N_{O}}$, where $\mathbf{n_i} \in \mathbb{R}^3$ are 3-dimensional positions $(x_i, y_i, z_i)$, and where $N_O$ is the number of nodes for the object (Assumption \textbf{A0}, Sec.~\ref{subsec:assumptions}). The state of the deformable object is fully described by the set of positions. By contrast, the trajectory transformation strategy employed by \MimicGen assumes that there is a single canonical coordinate frame for each object, and each object is fully described by the pose (position and rotation) of that coordinate frame. Such a coordinate frame is ill-defined for deformable objects, and even if such a frame could be assigned (for example to a particular node), it would not fully describe the state of the deformable object, and would violate the invariance assumption made by \MimicGen. Unlike rigid bodies, deformable objects exhibit continuous and high-dimensional configurations characterized by local deformations and complex material properties. As a result, they require more expressive, nonlinear models that can capture dynamic shape changes. In the following section, we discuss non-rigid registration, which provides a means to compare the configurations of deformable objects, and forms the basis for \sysName's data generation strategy.

\subsection{Non-Rigid Registration}
\label{subsec:nonrigid}

Consider a deformable object in two different configurations $O_1 = \{\mathbf{a_i}\}_{i=1}^{N}$ and $O_2 = \{\mathbf{b_i}\}_{i=1}^{N}$, where $\mathbf{a}, \mathbf{b} \in \mathbb{R}^3$ are position vectors. For example, this could be a towel that is crumpled in two different ways. Non-rigid registration finds a smooth function $\mathbf{f} : \mathbb{R}^3 \rightarrow \mathbb{R}^3$ that maps points from the first configuration to the second, see Fig.~\ref{fig:fig2} (Non-Rigid Registration). It does this by solving an optimization problem that minimizes a cost consisting of the point-wise distances between $\mathbf{f(a_i)}$ and $\mathbf{b_i}$ and a regularization term to encourage smoothness (see~\cite{schulman2016learning} for more details). Note that in general, non-rigid registration does not require the number of points in each configuration to be equal, nor does it require point-wise correspondences to be known beforehand~\cite{chui2003new, schulman2016learning}. In the next section, we describe how non-rigid registration can be used in the data generation process -- namely how the costs of the non-rigid registration optimization can be used for source demonstration selection, and how the resultant continuous deformation field $\mathbf{f}(\cdot)$ can be used to warp source demonstrations for data generation.

\begin{figure*}
    \centering
    \includegraphics[width=0.95\linewidth]{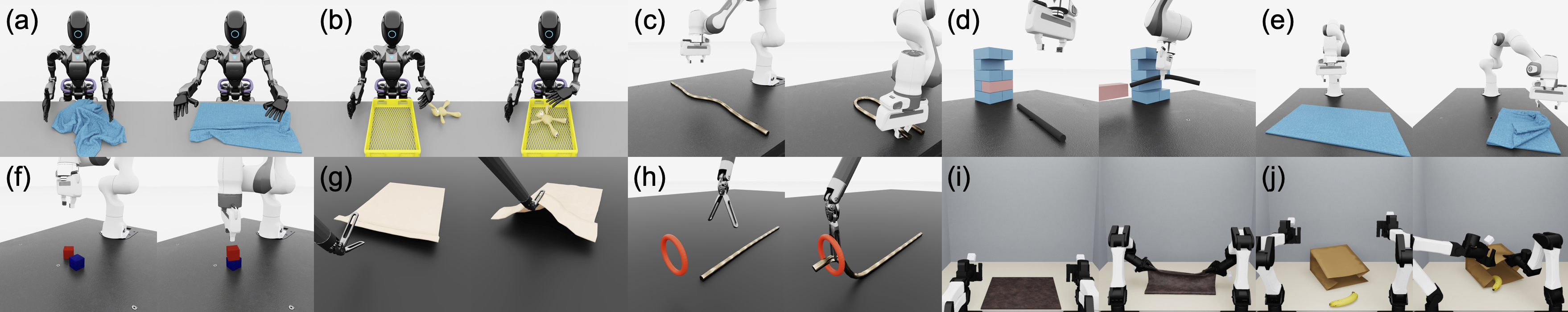}
    \caption{\textbf{Simulation Tasks.}
    \sysName is used to generate new demonstrations across 10 challenging tasks involving 4 distinct robot embodiments: (a-b) GR1 humanoid, (c-f) Franka arm, (g-h) dVRK surgical robot, and (i-j) bimanual YAM arms. These tasks demand high precision and fine-grained manipulation to be successfully executed.}
    \label{fig:fig3}
\end{figure*}

\subsection{Data Generation Mechanism}
\label{subsec:datagen_mechanism}

We now describe how we can leverage non-rigid registration during data generation. As described in Sec.~\ref{subsec:problem}, the main step during data generation is to select source demonstrations $\tau \in \mathcal{D}_{\text{src}}$ and adapt them to produce and execute a new robot trajectory $\tau'$.

Like \MimicGen, we start with source demonstrations $\mathcal{D}_{\text{src}}$ that are parsed into contiguous object-centric manipulation segments $\{\tau_i\}_{i=1}^M$, each of which corresponds to a subtask $S_i(o_i)$. Recall that each segment is a sequence of end-effector control poses $\tau_i = (T^{C_0}_W, T^{C_1}_W, ..., T^{C_K}_W)$ where $W$ is the world reference frame, and that this segmentation can be done autonomously with heuristics or using human annotation~\cite{mandlekar2023mimicgen}. To generate a demonstration in a novel scene, we proceed subtask by subtask.

\textbf{Source Demonstration Selection.}
For each subtask, we must first select a source demonstration segment from the set of segments for the current subtask. To do so, we observe the object configuration $O'_i = \{\mathbf{v_j}\}_{j=1}^{N_{O_{i}}}$ corresponding to the object for the current subtask. Next, we compare it against the relevant object configurations at the start of each source segment by running non-rigid registration and comparing the cost achieved by each optimization problem. This is an analog to the nearest-neighbor source segment selection strategy from \MimicGen~\cite{mandlekar2023mimicgen}, which selects source demos based on object pose distances between the new scene and the source demonstrations. Prior work~\cite{mandlekar2023mimicgen, robocasa2024} has found that intelligent source demonstration selection can be crucial for improving data generation quality -- this problem is exacerbated for deformable objects, since they have continuous, high-dimensional configuration spaces.

\textbf{Trajectory Adaptation.}
Next, the selected source demonstration segment $\tau_i = (T^{C_0}_W, T^{C_1}_W, ..., T^{C_K}_W)$ must be adapted to the current scene. To do so, we run non-rigid registration between the object configuration $O_i$ at the start of the source demonstration segment and the one in the current scene $O'_i$ and obtain a continuous deformation field $\mathbf{f(\cdot)}$. Next, each pose $T_t = (p_t, R_t)$ in the source segment can be transformed using the field~\cite{schulman2016learning}:
\[
p_t \rightarrow \mathbf{f}(p_t), \quad R_t \rightarrow \mathrm{orth}(\mathbf{J_f}(p_t) R_t)
\]
where $\mathbf{J_f}(p_t)$ is the Jacobian of $\mathbf{f}$ evaluated at $p_t$, and $\mathrm{orth}(\cdot)$ orthonormalizes the resulting matrix to yield a valid rotation. This transformation preserves the local spatial relationship between the end-effector and the deformable object as it deforms, Fig. \ref{fig:fig2} (Transform Segment). As in \MimicGen, a linear interpolation segment is added to the transformed trajectory segment $\tau'$ to ensure a smooth transition from the robot’s current pose to the start of the warped trajectory. The transformed trajectory is then executed using the robot’s controller. The process of source demonstration selection and trajectory adaptation and execution is repeated for every subtask, and if the demonstration achieves task success, it is added to the dataset.

By design, \sysName enables data generation in manipulation settings that go beyond rigid-body assumptions, including tasks involving flexible materials and complex deformations. Furthermore, the deformable object representation is equivalent to a point cloud representation for objects, and rigid-body objects could also be represented in this manner, making \sysName a strict generalization of \MimicGen. In fact, \sysName can be applied to rigid-body tasks and to rigid-body object geometries that are substantially different from the source demonstrations, unlike \MimicGen.

\section{Experimental Results}

In this section, we describe the experimental setup (Sec.~\ref{subsec:setup}), introduce our suite of deformable manipulation tasks (Sec.~\ref{subsec:tasks}), present empirical evidence highlighting the capabilities of \sysName (Sec.~\ref{subsec:features}), and conduct a systematic analysis of the system (Sec.~\ref{subsec:analysis}).

\subsection{Experiment Setup}
\label{subsec:setup}

We use Apple Vision Pro to collect source human demonstrations in our task suite (described in Sec.~\ref{subsec:tasks}). Our teleoperation pipeline retargets human hand motions to either a parallel-jaw gripper or a dexterous robotic hand, depending on the target embodiment. For the Franka robot and the surgical robot, we collect relative end-effector poses and gripper actions. For the GR1 humanoid robot, we collect absolute wrist poses and finger joint positions. The bimanual YAM arms are controlled in joint space.

We collect one to three source human demonstrations per task and use \sysName to generate 1,000 demonstrations per task, sampling from a broader initial state distribution. This setting reduces the burden on the human operator by limiting data collection to simpler task variations, while allowing the more challenging variations to be generated by \sysName. Each resulting dataset is used to train visuomotor policies using two imitation learning approaches: BC-RNN-GMM~\cite{mandlekar2021matters} and Diffusion Policy~\cite{chi2023diffusion}. For evaluation, we follow the protocol established in prior work~\cite{mandlekar2021matters}: each experiment is run with three different random seeds, and we report the maximum policy success rate across seeds, unless otherwise specified.

\subsection{Simulation Tasks}
\label{subsec:tasks}

We introduce a suite of high-fidelity deformable object manipulation tasks (Fig. \ref{fig:fig3}) implemented in Isaac Lab~\cite{mittal2025isaac}. They encompass a range of deformable objects (stuffed animal, rope, tissue, towel) and manipulation behaviors (high-precision threading, dynamic whipping, folding, pick-and-place) across four robot embodiments (Franka and YAM robot arms, humanoid, surgical robot).

\noindent \textbf{Humanoid -- Towel Unfold.} A humanoid robot unfolds a crumpled towel by executing a sequence of movements to spread it flat. The task is considered successful when the towel is laid out smoothly and the humanoid arms are retracted above the table.

\noindent \textbf{Humanoid -- Teddy.} A humanoid robot grasps the teddy plush toy and places it in a basket. The task is considered successful when the teddy bear is placed inside the basket, and the humanoid's left fingers are open and positioned above the basket.

\noindent \textbf{Franka -- Rope Manipulation.} The Franka robot performs a rope manipulation task by shaping a rope into a "U" configuration, starting from a randomly initialized state. The task is considered successful when the two ends of the rope are positioned close together.

\noindent \textbf{Franka -- Jenga.} The Franka robot performs dynamic whipping to remove a pink Jenga block from a Jenga tower. Both the tower configuration and the whip's initial state are randomized. The pink Jenga block is constrained to movement within the X-Y plane. The task is considered successful when the pink block is fully removed from the tower.

\noindent \textbf{Franka -- Towel.} The Franka robot folds a flat towel in half. The task is considered successful when the towel is folded and the robot arm is retracted above the table.

\noindent \textbf{Franka -- Rigid Cube Stack.} The Franka robot grasps a cube and places it on top of another cube. This task showcases the applicability of \sysName to manipulation that does not involve deformable objects.

\noindent \textbf{Surgical -- Tissue Manipulation.} A piece of soft tissue is fixed at two endpoints. The surgical robot uses forceps-style grippers to grasp the tissue and retract it upward.

\noindent \textbf{Surgical -- Threading.} The surgical robot grasps a soft thread and passes it through a ring. The task is considered successful if the thread passes through the ring.

\noindent \textbf{YAM -- Towel.} The bimanual YAM system folds a towel in half. The task is considered successful when the towel is folded and both arms have retracted.

\noindent \textbf{YAM -- Bag Loading.} The right YAM arm first opens a shopping bag, after which the left arm grasps and places a banana inside. The task is considered successful when the banana is placed in the bag and satisfies the specified placement threshold.

\subsection{\sysName Features}
\label{subsec:features}

\textbf{\sysName significantly reduces the burden of data collection for deformable object manipulation tasks.} Collecting robot data for deformable object manipulation is particularly challenging, as it requires fine-grained coordination between robot arms and, in the case of dexterous hands, precise grasping of soft objects which significantly increases operator burden. To address this, we collect only a small number of high-quality teleoperated demonstrations using the Apple Vision Pro. \sysName then leverages these seed demonstrations to generate large-scale synthetic datasets in simulation, covering a broader range of initial conditions and object states. The generation pipeline achieves success rates ranging from $70\%$ to $100\%$ across tasks. This approach enables efficient scaling of data collection while preserving task-relevant diversity, substantially reducing the human effort required to train performant visuomotor policies for deformable object manipulation tasks.

\textbf{\sysName improves policy performance relative to training only on source demonstrations.} As shown in Table \ref{tab:table1}, policies trained on \sysName-generated data consistently outperform those trained solely on human-collected demonstrations -- improvements range from $25\%$ to $97\%$. By using a small set of human teleoperated trajectories, \sysName enables more robust and effective policy learning, significantly reducing the reliance on costly and time-consuming human data collection.

\textbf{\sysName is applicable to a wide range of robots and object manipulation tasks.} 
\sysName successfully generates demonstrations across our entire task suite, which consists of four distinct platforms: GR1 humanoid, Franka Panda arm, surgical robot, and bimanual YAM arms, and includes diverse kinds of manipulation such as pick-and-place of a teddy bear, towel manipulation, dynamic whipping of a Jenga tower, surgical tissue manipulation, and high-precision threading. It even works for rigid object manipulation (Franka -- Rigid Cube), showing that it is a strict generalization of \MimicGen.

\begin{figure*}
    \centering
    \includegraphics[width=\linewidth]{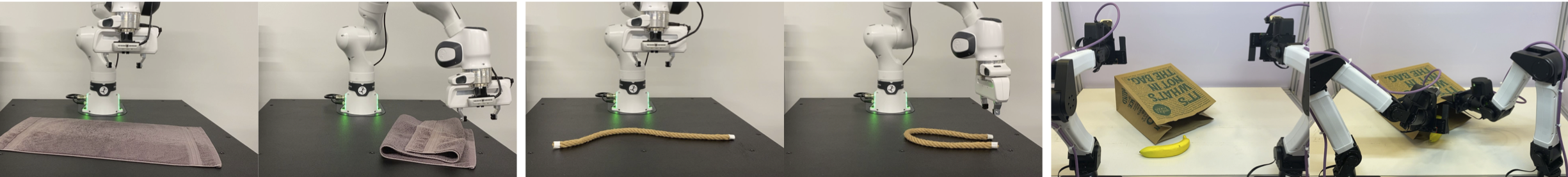}
    \caption{\textbf{Real-World Deployment.}
    Example rollouts of real-world deformable manipulation tasks using policies trained on \sysName-generated datasets.}
    \label{fig:fig4}
\end{figure*}

\begin{table}[]
\centering
\resizebox{\columnwidth}{!}{%
\begin{tabular}{l|c|c|c}
\toprule
\rowcolor[HTML]{CBCEFB} 
Task & \begin{tabular}[c]{@{}c@{}}Source Demo\\ BC-RNN-GMM\end{tabular} & \begin{tabular}[c]{@{}c@{}}Generated Demo\\ BC-RNN-GMM\end{tabular} & \begin{tabular}[c]{@{}c@{}}Generated Demo\\ Diffusion Policy\end{tabular} \\
\midrule
\midrule
Humanoid - Teddy&0.0 ± 0.0&32.0 ± 3.3&42.0 ± 2.0\\
\rowcolor[HTML]{EFEFEF} 
Humanoid - Towel&1.3 ± 1.9&50.7 ± 1.9&56.0 ± 5.7\\
Franka - Rope&2.0 ± 2.0&99.3 ± 0.9&100.0 ± 0.0\\
\rowcolor[HTML]{EFEFEF} 
Franka - Jenga&4.0 ± 3.3&89.3 ± 15.1&80.0 ± 3.3\\
Franka - Towel&0.0 ± 0.0&78.7 ± 6.8&70.7 ± 6.8\\
\rowcolor[HTML]{EFEFEF} 
Franka - Rigid Cube&24.0 ± 4.0&90.7 ± 5.0&50.7 ± 6.8\\
Surgical - Tissue&56.0 ± 5.7&81.3 ± 12.4&94.7 ± 1.9\\
\rowcolor[HTML]{EFEFEF} 
Surgical - Threading&5.3 ± 1.9&98.7 ± 1.9&58.7 ± 1.9\\
YAM - Towel&4.0 ± 3.3&13.3 ± 3.8&52.0 ± 18.2\\
\rowcolor[HTML]{EFEFEF}
YAM - Bag Loading&12.0 ± 6.5&14.7 ± 5.0&29.3 ± 5.0\\
\bottomrule
\end{tabular}%
}
\caption{\textbf{Policy Performance on Source and Generated Datasets.}
Success rates of visuomotor policies trained on source demonstrations and on \sysName-generated datasets. Success rates reported as the maximum over three training seeds.}
\label{tab:table1}
\end{table}

\subsection{\sysName Analysis}
\label{subsec:analysis}

\textbf{How does \sysName compare with \MimicGen?} We evaluate \MimicGen on the task of \textbf{Franka -- Rope Manipulation} and compare its data generation performance with that of \sysName. In this experiment, a single demonstration is collected in which a straight rope is manipulated into a U-shape. For \MimicGen, the object reference frame is centered at the midpoint of the rope, with the X-axis aligned along its length. During data generation, one half of the rope is kept stationary while the other half is randomized to produce a diverse distribution of initial rope configurations. \MimicGen successfully generates only 4 out of 50 demonstrations, primarily in cases where the free end of the rope closely aligns with the configurations present in the source demonstration. In comparison, \sysName achieves 49 out of 50 successful demonstrations, significantly outperforming \MimicGen. Since \MimicGen succeeds only on a narrow subset of configurations seen in the source data, policies trained on this data are unlikely to generalize to novel rope configurations.

\textbf{How does dataset size affect policy learning?} We train visuomotor policies on datasets of sizes 50, 250, 500, and 750, subsampled from the main generated dataset for each task. Table \ref{tab:table2} illustrates the effect of dataset size on policy performance. We observe that success rates generally increase with dataset size. This highlights the importance of larger datasets for training performant policies and underscores the value of scalable data generation without human supervision for deformable object manipulation tasks that require high levels of dexterity and dynamic control.

\begin{table}[]
\centering
\resizebox{\columnwidth}{!}{%
\begin{tabular}{l|c|c|c|c}
\toprule
\rowcolor[HTML]{CBCEFB} 
Task & 50 Demos & 250 Demos & 500 Demos & 750 Demos\\
\midrule
\midrule
Humanoid - Teddy&24.0 ± 0.0&26.0 ± 10.0&44.0 ± 0.0&44.0 ± 0.0\\
\rowcolor[HTML]{EFEFEF} 
Humanoid - Towel&61.3 ± 6.8&70.7 ± 6.8&64.0 ± 5.7&64.0 ± 9.8\\
Franka - Rope&82.7 ± 5.0&100.0 ± 0.0&98.7 ± 1.9&93.3 ± 6.8\\
\rowcolor[HTML]{EFEFEF} 
Franka - Jenga&53.3 ± 8.2&77.3 ± 6.8&93.3 ± 5.0&84.0 ± 15.0\\
Franka - Towel&81.3 ± 5.0&76.0 ± 3.3&74.7 ± 5.0&84.0 ± 3.3\\
\rowcolor[HTML]{EFEFEF} 
Franka - Rigid Cube&42.0 ± 2.0&68.0 ± 6.5&82.7 ± 3.8&84.0 ± 5.7\\
Surgical - Tissue&69.3 ± 15.1&56.0 ± 13.1&84.0 ± 5.7&82.7 ± 5.0\\
\rowcolor[HTML]{EFEFEF} 
Surgical - Threading&56.0 ± 8.0&84.0 ± 3.3&98.7 ± 1.9&96.0 ± 3.3\\
YAM - Towel&8.0 ± 0.0&12.0 ± 3.3&9.3 ± 1.9&17.3 ± 10.5\\
\rowcolor[HTML]{EFEFEF}
YAM - Bag Loading&6.7 ± 3.8&20.0 ± 9.8&17.3 ± 1.9&17.3 ± 1.9\\
\bottomrule
\end{tabular}%
}
\caption{\textbf{Dataset Size Comparison.}
Success rates of visuomotor policies trained on subsamples of each task’s generated dataset. Success rates reported as the maximum over three training seeds.}
\label{tab:table2}
\end{table}

\textbf{Replay-based mechanisms vs. training visuomotor policies.} Previous works have explored the use of scene registration and trajectory transfer as policies for deformable object manipulation \cite{schulman2016learning, schulman2013case}. These approaches typically rely on point clouds from depth sensors to register objects from a source demonstration to a new context. However, point clouds can be noisy, which negatively impacts registration accuracy and the overall success rate of data generation.

In \sysName, we instead leverage ground-truth nodal information provided by the soft-body simulator to perform precise scene registration, enabling large-scale dataset generation. While our generation pipeline achieves high success rates, we use the generated data to train visuomotor policies through imitation learning. For our simulation results in Tables~\ref{tab:table1} and \ref{tab:table2}, these policies learn to manipulate deformable objects directly from raw image input, thereby bypassing explicit registration at inference time.

\subsection{Real-World Evaluation}

We evaluate policies trained on datasets generated by \sysName in a real-world setup on three deformable manipulation tasks shown in Fig.~\ref{fig:fig4}. For each task, we generate 1,000 synthetic demonstrations using \sysName from a single human teleoperated demonstration. We consider three training settings: real-only training, where policies are trained on 30 real-world demonstrations; zero-shot sim-to-real transfer, where policies are trained only on the 1,000 simulated demonstrations; and sim–real co-training~\cite{maddukuri2025sim}, where policies are trained jointly on the 30 real demonstrations and the 1,000 \sysName-generated simulated demonstrations.

For real-world evaluation, we leverage Point Bridge~\cite{haldar2026point}, which uses unified, domain-agnostic point-based representations to bridge the sim-to-real gap. During simulation data generation, we extract point clouds of deformable objects using ground-truth masks and depth maps. At deployment, Point Bridge’s VLM-guided pipeline extracts task-relevant object points from RGB-D camera observations. Policies are trained and deployed on these point-based observations and output actions, without using non-rigid registration as an explicit online controller.

\begin{table}[]
\centering
\resizebox{\columnwidth}{!}{%
\begin{tabular}{l|c|c|c}
\toprule
\rowcolor[HTML]{CBCEFB} 
Task & \begin{tabular}[c]{@{}c@{}}Real \\ 30 Demos\end{tabular} & \begin{tabular}[c]{@{}c@{}}Zero-shot Sim \\ 1,000 Demos\end{tabular} & \begin{tabular}[c]{@{}c@{}}Sim-Real Co-Train \\ 1,000 + 30 Demos\end{tabular} \\
\midrule
\midrule
Franka - Towel&76.6&70.0&76.6\\
\rowcolor[HTML]{EFEFEF} 
Franka - Rope&46.7&33.3&76.6\\
YAM - Bag Loading&33.3&63.3&93.3\\
\bottomrule
\end{tabular}%
}
\caption{\textbf{Real-World Deployment Results.}
Performance comparison across three training settings: real-only (30 demonstrations), zero-shot sim-to-real (1,000 simulation demonstrations), and sim–real co-training (1,000 simulation + 30 real demonstrations). Results show that large-scale simulation enables effective real-world transfer, while \sysName-generated data improves performance under limited real-world supervision.}
\label{tab:table3}
\end{table}

\textbf{Policies trained on \sysName-generated data achieve zero-shot sim-to-real transfer and are further improved with sim-real co-training.}
Table \ref{tab:table3} reports results for three settings: real-only training, zero-shot sim-to-real transfer, and sim–real co-training. The results show that policies trained with large-scale simulation data can successfully transfer to real-world tasks. Furthermore, \sysName-generated data improves the performance of policies trained with limited real-world demonstrations.

\section{Conclusion}

We introduce \sysName, an automated data generation pipeline that synthesizes large-scale datasets for deformable object manipulation from a small number of human demonstrations, and a suite of high-fidelity simulation environments that encompasses a wide range of deformable objects and manipulation behaviors across four different robot embodiments. We apply \sysName to generate datasets across the task suite and show that policies trained on data generated by \sysName achieve strong performance across diverse task distributions, including tasks requiring precision and dynamic behavior. \sysName assumes a fixed sequence of object-centric subtasks. Many real-world deformable manipulation tasks are less structured and may require multiple attempts or conditional transitions. Extending \sysName to support flexible task structures is a promising direction for future work. We hope that our simulation environments, data generation pipeline, and datasets greatly reduce the barrier for practitioners to study deformable object manipulation, especially in the context of imitation learning and robot foundation models.

\vspace{-5pt}

\section*{Acknowledgment}

We thank Simon Schirm and Siddhant Haldar for their valuable discussions and insights.

\vspace{-5pt}

\bibliographystyle{IEEEtran}
\bibliography{root}

@inproceedings{o2024open,
  title={Open x-embodiment: Robotic learning datasets and rt-x models: Open x-embodiment collaboration 0},
  author={O’Neill, Abby and Rehman, Abdul and Maddukuri, Abhiram and Gupta, Abhishek and Padalkar, Abhishek and Lee, Abraham and Pooley, Acorn and Gupta, Agrim and Mandlekar, Ajay and Jain, Ajinkya and others},
  booktitle={2024 IEEE Int'l Conf on Robotics and Automation (ICRA)},
  ign-pages={6892--6903},
  year={2024},
  ign-organization={IEEE}
}

@article{mandlekar2021matters,
  title={What matters in learning from offline human demonstrations for robot manipulation},
  author={Mandlekar, Ajay and Xu, Danfei and Wong, Josiah and Nasiriany, Soroush and Wang, Chen and Kulkarni, Rohun and Fei-Fei, Li and Savarese, Silvio and Zhu, Yuke and Mart{\'\i}n-Mart{\'\i}n, Roberto},
  journal={arXiv preprint arXiv:2108.03298},
  year={2021}
}

@article{brohan2022rt,
  title={Rt-1: Robotics transformer for real-world control at scale},
  author={Brohan, Anthony and Brown, Noah and Carbajal, Justice and Chebotar, Yevgen and Dabis, Joseph and Finn, Chelsea and Gopalakrishnan, Keerthana and Hausman, Karol and Herzog, Alex and Hsu, Jasmine and others},
  journal={arXiv preprint arXiv:2212.06817},
  year={2022}
}

@article{khazatsky2024droid,
  title={Droid: A large-scale in-the-wild robot manipulation dataset},
  author={Khazatsky, Alexander and Pertsch, Karl and Nair, Suraj and Balakrishna, Ashwin and Dasari, Sudeep and Karamcheti, Siddharth and Nasiriany, Soroush and Srirama, Mohan Kumar and Chen, Lawrence Yunliang and Ellis, Kirsty and others},
  journal={arXiv preprint arXiv:2403.12945},
  year={2024}
}

@inproceedings{mandlekar2018roboturk,
  title={Roboturk: A crowdsourcing platform for robotic skill learning through imitation},
  author={Mandlekar, Ajay and Zhu, Yuke and Garg, Animesh and Booher, Jonathan and Spero, Max and Tung, Albert and Gao, Julian and Emmons, John and Gupta, Anchit and Orbay, Emre and others},
  booktitle={Conf on Robot Learning},
  ign-pages={879--893},
  year={2018},
  ign-organization={PMLR}
}

@article{mandlekar2023mimicgen,
  title={Mimicgen: A data generation system for scalable robot learning using human demonstrations},
  author={Mandlekar, Ajay and Nasiriany, Soroush and Wen, Bowen and Akinola, Iretiayo and Narang, Yashraj and Fan, Linxi and Zhu, Yuke and Fox, Dieter},
  journal={arXiv preprint arXiv:2310.17596},
  year={2023}
}

@article{jiang2024dexmimicgen,
  title={Dexmimicgen: Automated data generation for bimanual dexterous manipulation via imitation learning},
  author={Jiang, Zhenyu and Xie, Yuqi and Lin, Kevin and Xu, Zhenjia and Wan, Weikang and Mandlekar, Ajay and Fan, Linxi and Zhu, Yuke},
  journal={arXiv preprint arXiv:2410.24185},
  year={2024}
}

@article{garrett2024skillmimicgen,
  title={Skillmimicgen: Automated demonstration generation for efficient skill learning and deployment},
  author={Garrett, Caelan and Mandlekar, Ajay and Wen, Bowen and Fox, Dieter},
  journal={arXiv preprint arXiv:2410.18907},
  year={2024}
}

@inproceedings{schulman2016learning,
  title={Learning from demonstrations through the use of non-rigid registration},
  author={Schulman, John and Ho, Jonathan and Lee, Cameron and Abbeel, Pieter},
  booktitle={Robotics Research: The 16th Int'l Symposium ISRR},
  ign-pages={339--354},
  year={2016},
  ign-organization={Springer}
}

@inproceedings{schulman2013case,
  title={A case study of trajectory transfer through non-rigid registration for a simplified suturing scenario},
  author={Schulman, John and Gupta, Ankush and Venkatesan, Sibi and Tayson-Frederick, Mallory and Abbeel, Pieter},
  booktitle={2013 IEEE/RSJ Int'l Conf on Intelligent Robots and Systems},
  ign-pages={4111--4117},
  year={2013},
  ign-organization={IEEE}
}

@article{mittal2025isaac,
  title={Isaac lab: A gpu-accelerated simulation framework for multi-modal robot learning},
  author={Mittal, Mayank and Roth, Pascal and Tigue, James and Richard, Antoine and Zhang, Octi and Du, Peter and Serrano-Munoz, Antonio and Yao, Xinjie and Zurbr{\"u}gg, Ren{\'e} and Rudin, Nikita and others},
  journal={arXiv preprint arXiv:2511.04831},
  year={2025}
}

@article{chi2023diffusion,
  title={Diffusion policy: Visuomotor policy learning via action diffusion},
  author={Chi, Cheng and Xu, Zhenjia and Feng, Siyuan and Cousineau, Eric and Du, Yilun and Burchfiel, Benjamin and Tedrake, Russ and Song, Shuran},
  journal={The Int'l Journal of Robotics Research},
  ign-pages={02783649241273668},
  year={2023},
  ign-publisher={SAGE Publications Sage UK: London, England}
}

@article{brohan2023rt,
  title={Rt-2: Vision-language-action models transfer web knowledge to robotic control},
  author={Brohan, Anthony and Brown, Noah and Carbajal, Justice and Chebotar, Yevgen and Chen, Xi and Choromanski, Krzysztof and Ding, Tianli and Driess, Danny and Dubey, Avinava and Finn, Chelsea and others},
  journal={arXiv preprint arXiv:2307.15818},
  year={2023}
}

@article{kim2024openvla,
  title={{OpenVLA}: An open-source vision-language-action model},
  author={Kim, Moo Jin and Pertsch, Karl and Karamcheti, Siddharth and Xiao, Ted and Balakrishna, Ashwin and Nair, Suraj and Rafailov, Rafael and Foster, Ethan and Lam, Grace and Sanketi, Pannag and others},
  journal={arXiv preprint arXiv:2406.09246},
  year={2024}
}

@article{black2024pi_0,
  title={$\pi_0 $: A Vision-Language-Action Flow Model for General Robot Control},
  author={Black, Kevin and Brown, Noah and Driess, Danny and Esmail, Adnan and Equi, Michael and Finn, Chelsea and Fusai, Niccolo and Groom, Lachy and Hausman, Karol and Ichter, Brian and others},
  journal={arXiv preprint arXiv:2410.24164},
  year={2024}
}

@article{bjorck2025gr00t,
  title={GR00T N1: An Open Foundation Model for Generalist Humanoid Robots},
  author={NVIDIA and Bjorck, Johan and Casta{\~n}eda, Fernando and Cherniadev, Nikita and Da, Xingye and Ding, Runyu and Fan, Linxi and Fang, Yu and Fox, Dieter and Hu, Fengyuan and Huang, Spencer and others},
  journal={arXiv preprint arXiv:2503.14734},
  year={2025}
}

@inproceedings{wang2023robogen,
  title={RoboGen: Towards Unleashing Infinite Data for Automated Robot Learning via Generative Simulation},
  author={Wang, Yufei and Xian, Zhou and Chen, Feng and Wang, Tsun-Hsuan and Wang, Yian and Fragkiadaki, Katerina and Erickson, Zackory and Held, David and Gan, Chuang},
  booktitle={{Forty-first Int'l Conf on Machine Learning}},
  year={2023},
}

@inproceedings{robocasa2024,
  title={RoboCasa: Large-Scale Simulation of Everyday Tasks for Generalist Robots},
  author={Soroush Nasiriany and Abhiram Maddukuri and Lance Zhang and Adeet Parikh and Aaron Lo and Abhishek Joshi and Ajay Mandlekar and Yuke Zhu},
  booktitle={Robotics: Science and Systems (RSS)},
  year={2024}
}

@inproceedings{todorov2012mujoco,
  title={Mujoco: A physics engine for model-based control},
  author={Todorov, Emanuel and Erez, Tom and Tassa, Yuval},
  booktitle={IEEE/RSJ Int'l Conf on Intelligent Robots and Systems},
  ign-pages={5026--5033},
  year={2012},
}

@inproceedings{dalal2023imitating,
  title={Imitating Task and Motion Planning with Visuomotor Transformers},
  author={Dalal, Murtaza and Mandlekar, Ajay and Garrett, Caelan Reed and Handa, Ankur and Salakhutdinov, Ruslan and Fox, Dieter},
  booktitle={{Conf on Robot Learning}},
  ign-pages={2565--2593},
  year={2023},
  ign-organization={PMLR}
}

@article{maddukuri2025sim,
  title={Sim-and-real co-training: A simple recipe for vision-based robotic manipulation},
  author={Maddukuri, Abhiram and Jiang, Zhenyu and Chen, Lawrence Yunliang and Nasiriany, Soroush and Xie, Yuqi and Fang, Yu and Huang, Wenqi and Wang, Zu and Xu, Zhenjia and Chernyadev, Nikita and others},
  journal={arXiv preprint arXiv:2503.24361},
  year={2025}
}

@article{wei2025empirical,
  title={Empirical analysis of sim-and-real cotraining of diffusion policies for planar pushing from pixels},
  author={Wei, Adam and Agarwal, Abhinav and Chen, Boyuan and Bosworth, Rohan and Pfaff, Nicholas and Tedrake, Russ},
  journal={arXiv preprint arXiv:2503.22634},
  year={2025}
}

@inproceedings{zeng2020transporter,
  title={Transporter networks: Rearranging the visual world for robotic manipulation},
  author={Zeng, Andy and Florence, Pete and Tompson, Jonathan and Welker, Stefan and Chien, Jonathan and Attarian, Maria and Armstrong, Travis and Krasin, Ivan and Duong, Dan and Sindhwani, Vikas and others},
  booktitle={Conf on Robot Learning},
  ign-pages={726--747},
  year={2021},
  ign-organization={PMLR}
}

@INPROCEEDINGS{zhao2023learning, 
    AUTHOR    = {Tony Z. Zhao AND Vikash Kumar AND Sergey Levine AND Chelsea Finn}, 
    TITLE     = {{Learning Fine-Grained Bimanual Manipulation with Low-Cost Hardware}}, 
    BOOKTITLE = {Robotics: Science and Systems}, 
    YEAR      = {2023}, 
    ADDRESS   = {Daegu, Republic of Korea}, 
    ign-month     = {July}, 
    ign-doi       = {10.15607/RSS.2023.XIX.016} 
}

@INPROCEEDINGS{ebert2021bridge, 
    AUTHOR    = {Frederik Ebert AND Yanlai Yang AND Karl Schmeckpeper AND Bernadette Bucher AND Georgios Georgakis AND Kostas Daniilidis AND Chelsea Finn AND Sergey Levine}, 
    TITLE     = {{Bridge Data: Boosting Generalization of Robotic Skills with Cross-Domain Datasets}}, 
    BOOKTITLE = {Robotics: Science and Systems}, 
    YEAR      = {2022}, 
    ign-month     = {June}, 
    ign-doi       = {10.15607/RSS.2022.XVIII.063} 
}

@article{james2020rlbench,
  title={Rlbench: The robot learning benchmark \& learning environment},
  author={James, Stephen and Ma, Zicong and Arrojo, David Rovick and Davison, Andrew J},
  journal={IEEE Robotics and Automation Letters},
  volume={5},
  ign-number={2},
  ign-pages={3019--3026},
  year={2020},
  ign-publisher={IEEE}
}

@inproceedings{kostrikov2020image,
  title={Image augmentation is all you need: Regularizing deep reinforcement learning from pixels},
  author={Yarats, Denis and Kostrikov, Ilya and Fergus, Rob},
  booktitle={Int'l Conf on learning representations},
  year={2021}
}

@article{young2020visual,
  title={Visual imitation made easy},
  author={Young, Sarah and Gandhi, Dhiraj and Tulsiani, Shubham and Gupta, Abhinav and Abbeel, Pieter and Pinto, Lerrel},
  journal={arXiv e-prints},
  ign-pages={arXiv--2008},
  year={2020}
}

@article{yu2023scaling,
  title={Scaling robot learning with semantically imagined experience},
  author={Yu, Tianhe and Xiao, Ted and Stone, Austin and Tompson, Jonathan and Brohan, Anthony and Wang, Su and Singh, Jaspiar and Tan, Clayton and Peralta, Jodilyn and Ichter, Brian and others},
  journal={arXiv preprint arXiv:2302.11550},
  year={2023}
}

@article{chen2023genaug,
  title={GenAug: Retargeting behaviors to unseen situations via Generative Augmentation},
  author={Chen, Zoey and Kiami, Sho and Gupta, Abhishek and Kumar, Vikash},
  journal={arXiv preprint arXiv:2302.06671},
  year={2023}
}

@inproceedings{pomerleau1989alvinn,
  title={Alvinn: An autonomous land vehicle in a neural network},
  author={Pomerleau, Dean A},
  booktitle={Advances in neural information processing systems},
  ign-pages={305--313},
  year={1989}
}

@article{schaal1999imitation,
  title={Is imitation learning the route to humanoid robots?},
  author={Schaal, Stefan},
  journal={Trends in cognitive sciences},
  volume={3},
  ign-number={6},
  ign-pages={233--242},
  year={1999},
  ign-publisher={Elsevier}
}

@article{Ijspeert2002MovementIW,
  title={Movement imitation with nonlinear dynamical systems in humanoid robots},
  author={Auke Jan Ijspeert and Jun Nakanishi and Stefan Schaal},
  journal={Proceedings 2002 IEEE Int'l Conf on Robotics and Automation},
  year={2002},
  volume={2},
  ign-pages={1398-1403 vol.2}
}

@inproceedings{Billard2008RobotPB,
  title={Robot Programming by Demonstration},
  author={Aude Billard and Sylvain Calinon and R{\"u}diger Dillmann and Stefan Schaal},
  booktitle={Springer Handbook of Robotics},
  year={2008}
}

@article{Calinon2010LearningAR,
  title={Learning and Reproduction of Gestures by Imitation},
  author={Sylvain Calinon and Florent D'halluin and Eric L. Sauser and Darwin G. Caldwell and Aude Billard},
  journal={IEEE Robotics and Automation Magazine},
  year={2010},
  volume={17},
  ign-pages={44-54}
}

@article{zhu2022domsurvey,
  title={Challenges and outlook in robotic manipulation of deformable objects},
  author={Zhu, Jihong and Cherubini, Andrea and Dune, Claire and Navarro-Alarcon, David and Alambeigi, Farshid and Berenson, Dmitry and Ficuciello, Fanny and Harada, Kensuke and Kober, Jens and Li, Xiang and others},
  journal={IEEE Robotics \& Automation Magazine},
  volume={29},
  number={3},
  pages={67--77},
  year={2022},
  publisher={IEEE}
}

@article{yin2021modeling,
  title={Modeling, learning, perception, and control methods for deformable object manipulation},
  author={Yin, Hang and Varava, Anastasia and Kragic, Danica},
  journal={Science Robotics},
  year={2021},
}

@inproceedings{feng2024pie,
  title={Pie-nerf: Physics-based interactive elastodynamics with nerf},
  author={Feng, Yutao and Shang, Yintong and Li, Xuan and Shao, Tianjia and Jiang, Chenfanfu and Yang, Yin},
  booktitle={IEEE/CVF Conf on Computer Vision and Pattern Recognition},
  year={2024}
}

@article{qiao2022neuphysics,
  title={Neuphysics: Editable neural geometry and physics from monocular videos},
  author={Qiao, Yi-Ling and Gao, Alexander and Lin, Ming},
  journal={Advances in Neural Information Processing Systems},
  year={2022}
}

@inproceedings{zhong2024reconstruction,
  title={Reconstruction and simulation of elastic objects with spring-mass 3d gaussians},
  author={Zhong, Licheng and Yu, Hong-Xing and Wu, Jiajun and Li, Yunzhu},
  booktitle={European Conf on Computer Vision},
  year={2024}
}

@article{shi2023robocook,
  title={Robocook: Long-horizon elasto-plastic object manipulation with diverse tools},
  author={Shi, Haochen and Xu, Huazhe and Clarke, Samuel and Li, Yunzhu and Wu, Jiajun},
  journal={arXiv preprint arXiv:2306.14447},
  year={2023}
}

@article{li2020causal,
  title={Causal discovery in physical systems from videos},
  author={Li, Yunzhu and Torralba, Antonio and Anandkumar, Anima and Fox, Dieter and Garg, Animesh},
  journal={Advances in Neural Information Processing Systems},
  year={2020}
}

@inproceedings{sanchez2020learning,
  title={Learning to simulate complex physics with graph networks},
  author={Sanchez-Gonzalez, Alvaro and Godwin, Jonathan and Pfaff, Tobias and Ying, Rex and Leskovec, Jure and Battaglia, Peter},
  booktitle={Int'l Conf on machine learning},
  year={2020}
}

@article{pitis2020counterfactual,
  title={Counterfactual data augmentation using locally factored dynamics},
  author={Pitis, Silviu and Creager, Elliot and Garg, Animesh},
  journal={Advances in Neural Information Processing Systems},
  year={2020}
}

@article{pitis2022mocoda,
  title={Mocoda: Model-based counterfactual data augmentation},
  author={Pitis, Silviu and Creager, Elliot and Mandlekar, Ajay and Garg, Animesh},
  journal={Advances in Neural Information Processing Systems},
  year={2022}
}

@article{chui2003new,
  title={A new point matching algorithm for non-rigid registration},
  author={Chui, Haili and Rangarajan, Anand},
  journal={Computer Vision and Image Understanding},
  volume={89},
  number={2-3},
  pages={114--141},
  year={2003},
  publisher={Elsevier}
}

@inproceedings{lee2014unifying,
  title={Unifying scene registration and trajectory optimization for learning from demonstrations with application to manipulation of deformable objects},
  author={Lee, Alex X and Huang, Sandy H and Hadfield-Menell, Dylan and Tzeng, Eric and Abbeel, Pieter},
  booktitle={2014 IEEE/RSJ International Conference on Intelligent Robots and Systems},
  pages={4402--4407},
  year={2014},
  organization={IEEE}
}

@article{haldar2026point,
  title={Point Bridge: 3D Representations for Cross Domain Policy Learning},
  author={Haldar, Siddhant and Johannsmeier, Lars and Pinto, Lerrel and Gupta, Abhishek and Fox, Dieter and Narang, Yashraj and Mandlekar, Ajay},
  journal={arXiv preprint arXiv:2601.16212},
  year={2026}
}

\end{document}